\def\paperID{14}
\def\confName{CVPR}
\def\confYear{2025}

\def\paperTitle{
Simple Visual Artifact Detection in Sora-Generated Videos
}

\def\authorBlock{
    Misora Sugiyama$^{\dag}$ \ \ Hirokatsu Kataoka$^{\ddag,\diamondsuit}$ \\
    \dag Joshigakuin Senior High School \\
    \ddag National Institute of Advanced Industrial Science and Technology (AIST) \\ $\diamondsuit$ Visual Geometry Group, University of Oxford \\
    {\tt\small \ misorasugiyama.eca@gmail.com}
}

\newif\ifreview 
\newif\ifarxiv 
\newif\ifcamera \newcommand{\cameraready}{\cameratrue}
\newif\ifrebuttal

\cameraready
\pdfoutput=1
\documentclass[10pt,twocolumn,letterpaper]{article}

\ifreview \usepackage[review]{cvpr} \fi
\ifarxiv \usepackage[pagenumbers]{cvpr} \fi
\ifrebuttal \usepackage[rebuttal]{cvpr} \fi
\ifcamera \usepackage{cvpr} \fi

\usepackage{graphicx}	
\usepackage{amsmath}	
\usepackage{amssymb}	
\usepackage{booktabs}
\usepackage{times}
\usepackage{microtype}
\usepackage{epsfig}
\usepackage{caption}
\usepackage{float}
\usepackage{placeins}
\usepackage{color, colortbl}
\usepackage{stfloats}
\usepackage{enumitem}
\usepackage{tabularx}
\usepackage{xstring}
\usepackage{multirow}
\usepackage{xspace}
\usepackage{url}
\usepackage{subcaption}
\usepackage{xcolor}
\usepackage[hang,flushmargin]{footmisc}
\usepackage{comment}

\ifcamera \usepackage[accsupp]{axessibility} \fi

% ...

\ifarxiv  \fi

\newcommand{\R}[1]{{%
    \textbf{%
        \ifstrequal{#1}{1}{\textcolor{red}{R#1}}{%
        \ifstrequal{#1}{2}{\textcolor{blue}{R#1}}{%
        \ifstrequal{#1}{3}{\textcolor{magenta}{R#1}}{%
        \ifstrequal{#1}{4}{\textcolor{teal}{R#1}}{%
                           \textcolor{cyan}{R#1}%
        }}}}%
    }%
}}

\newcommand{\hk}[1]{{\color{black}{#1}}}
\usepackage{xr-hyper}

\makeatletter
\newcommand*{\addFileDependency}[1]{
  \typeout{(#1)}
  \@addtofilelist{#1}
  \IfFileExists{#1}{}{\typeout{No file #1.}}
}

\makeatother

\definecolor{cvprblue}{rgb}{0.21,0.49,0.74}
\usepackage[pagebackref,breaklinks,colorlinks,allcolors=cvprblue]{hyperref}
\usepackage[capitalize]{cleveref}
\crefname{section}{Sec.}{Secs.}
\crefname{table}{Table}{Tables}
\crefname{figure}{Fig.}{Figs.}

\ifarxiv \crefname{appendix}{App.}{Apps.}
\else \crefname{appendix}{Suppl.}{Suppls.} \fi

\begin{document}
%% TITLE
\title{\paperTitle}
\author{\authorBlock}
\maketitle

\begin{abstract}
The December 2024 release of OpenAI’s Sora, a powerful video generation model driven by natural language prompts, highlights a growing convergence between large language models (LLMs) and video synthesis. As these multimodal systems evolve into video-enabled LLMs (VidLLMs), capable of interpreting, generating, and interacting with visual content, understanding their limitations and ensuring their safe deployment becomes essential. This study investigates visual artifacts frequently found and reported in Sora-generated videos, which can compromise quality, mislead viewers, or propagate disinformation. We propose a multi-label classification framework targeting four common artifact label types: label 1: boundary / edge defects, label 2: texture / noise issues, label 3: movement / joint anomalies, and label 4: object mismatches / disappearances. Using a dataset of 300 manually annotated frames extracted from 15 Sora-generated videos, we trained multiple 2D CNN architectures (ResNet-50, EfficientNet-B3 / B4, ViT-Base). The best-performing model trained by ResNet-50 achieved an average multi-label classification accuracy of 94.14\%. This work supports the broader development of VidLLMs by contributing to (1) the creation of datasets for video quality evaluation, (2) interpretable artifact-based analysis beyond language metrics, and (3) the identification of visual risks relevant to factuality and safety. 

\end{abstract}

% fig
\begin{figure*}[t]
    \centering
    \includegraphics[width=0.95\linewidth]{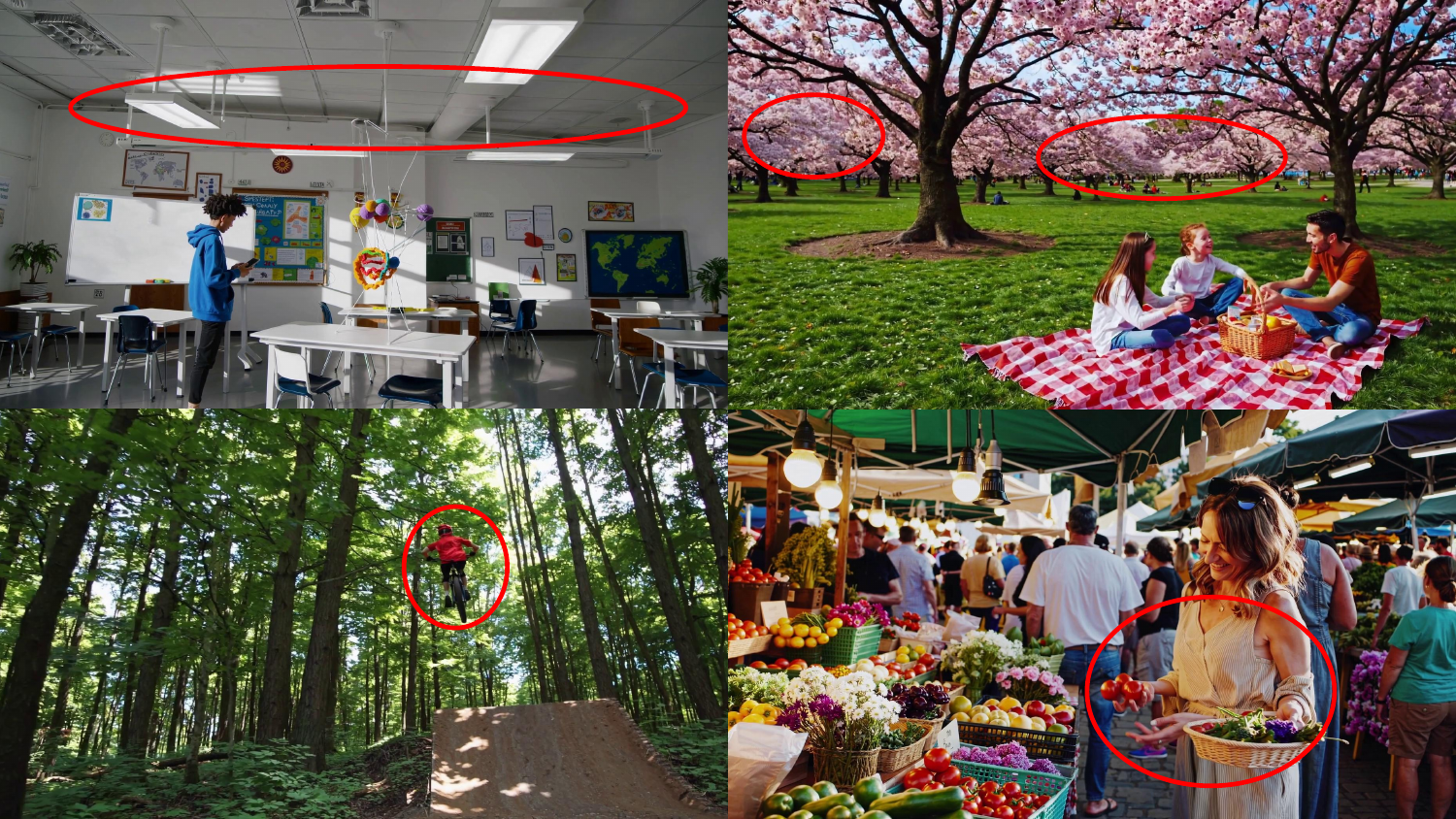}
    \caption{Examples of four artifact categories: top left “boundary/edge defects,” top right “texture/noise issues,” bottom left “motion/joint anomalies,” bottom right “object mismatches / disappearances”}
    \label{fig:dataset_pipline}
\end{figure*}

% From the CfP in CVPR 2025 Workshop on VideoLLM 
% Data Creation: Innovative strategies for leveraging web video data, advanced filtering techniques, synthetic data generation, long video understanding datasets and enriching datasets for video instruction tuning.
% Evaluation and Analysis: Robust evaluation frameworks for existing models, focusing on improving interpretability, deriving novel insights, and introducing new metrics and benchmarks for VidLLMs.
% Limitations, Risks and Safety: Addressing bias, fairness, and ethical challenges in VidLLMs, including factuality, hallucination, and safety concerns.

\section{Introduction}

Recent advances in generative AI have enabled large-scale language models (LLMs) to interface with multi-modal data, extending beyond text to images, audio, and video synthesis~\cite{liu2024sora, Yin_2024}. OpenAI’s Sora, publicly released in December 2024, represents a significant advancement in text-to-video generation, allowing the production of highly realistic and temporally coherent video content from simple natural language prompts~\cite{anantrasirichai2025artificialintelligencecreativeindustries, liu2024sora}. However, in addition to these opportunities, there are serious challenges in terms of trustworthiness, quality assurance, and safety. AI-generated videos often exhibit subtle yet significant artifacts that may alter visual coherence or mislead viewers~\cite{Banerjee2024,li2018exposing, rossler2019faceforensics++}. Such artifacts, ranging from edge discontinuities to implausible object deformations, are particularly concerning in applications involving factual storytelling, education, or the spread of information. As video generation systems increasingly integrate with LLMs to form video language models (VidLLMs), tools are urgently needed to detect and categorize these visual failures systematically~\cite{liu2025lavid, ma2024decof}.

Even as a social issue, there have been news stories designed to mislead for centuries. The advent of modern digital technology and social media platforms has dramatically increased the influence of fake news~\cite{allcott2017social, FakeNews_2018, tandoc2018defining} . For example, the 2016 US presidential election highlighted the unprecedented impact of online misinformation. It revealed that traditional fact-checking methods are inadequate to combat the vast and sophisticated misleading content enhanced by new AI technologies~\cite{guess2020exposure, thies2016face2face, vosoughi2018spread}. In this evolving landscape, deep faking techniques have garnered significant attention; early academic research, such as Face2Face project in 2016, demonstrated that real-time facial movements of an individual could be convincingly transferred to another person's video~\cite{suwajanakorn2017synthesizing}. Similarly, the "Synthesizing Obama" study in 2017 introduced the ability to map entirely new speech content onto existing footage of former President Obama, resulting in videos that made it appear that he was speaking what he was not actually speaking~\cite{GAN_2014}. 

Advances in deep neural networks, particularly generative adversarial networks (GANs), showed the potential for high-quality image generation. GANs have established a powerful paradigm based on adversarial training, where generator and discriminator networks are trained in opposition to each other to produce realistic images~\cite{karras2017progressive}. GANs have also further streamlined the process of creating deepfakes, making these techniques accessible to non-experts and exacerbating concerns about the authenticity of digital media~\cite{bond2021deep}. Although GANs boast high accuracy in image generation, they also present problems such as unstable training and mode collapse. Variational Autoencoders (VAEs) have attracted attention as an alternative method to solve these problems~\cite{bond2021deep, sakirin2023survey}.
Variational Autoencoders (VAEs) ~\cite{kingma2013auto} offer a complementary framework that leverages probabilistic inference by encoding data into a latent space and decoding it back to the original space. This approach provides smooth interpolation in the latent domain and facilitates efficient inference, making VAEs particularly attractive for scenarios where interpretable latent representations are valuable. More recently, diffusion models~\cite{ho2020denoising, nichol2021improved} have gained prominence due to their innovative strategy of gradually denoising random noise to generate high-quality images. This iterative refinement process has led to state-of-the-art performance in image synthesis tasks, and diffusion-based methods have also been adapted for conditional generation scenarios.

Integrating natural language processing with generative models has further propelled the field toward multimodal applications. Text-to-image generation, as illustrated by Ramesh et al.~\cite{ramesh2021zero}, leverages language inputs to guide the synthesis process, yielding visually coherent outputs that align with textual descriptions. Recent efforts in text-to-video generation have extended this paradigm and focused on capturing temporal dynamics. Notable examples include Make-A-Video ~\cite{singer2022make} and the large-scale pretraining approach described in CogVideo~\cite{hong2022cogvideolargescalepretrainingtexttovideo}. These represent promising steps toward generating temporally coherent video sequences directly from textual prompts. The evolution from GANs to VAEs and diffusion models has enriched the generation technology landscape and contributed to the increase in multimodal generation systems.

These techniques have pushed beyond traditional static image generation and paved the way for more advanced multimodal generation. OpenAI's Sora, a text-to-video generative AI model, aims to improve generation capabilities in complex multimodal environments. This research focuses on Sora-generated videos, analyzing ``artifacts'' from the AI's generative process. In this paper, we simply investigate a multi-label classification approach to identify four main artifact categories, label 1: boundary / edge defects, label 2: texture / noise issues, label 3: movement / joint anomalies, and  label 4: object mismatches / disappearances~\cite{chang2024matters}. Even using multiple 2D CNN architectures (ResNet-50, EfficientNet-B3 / B4, ViT-Base) on labeled artifact types (label 1 - 4), we contribute to broader efforts in safeguarding information integrity on digital platforms. This study works on simple visual artifact detection in AI-generated videos to improve reliability and safety. According to the fact from our experimental results, it has a great chance to effectively detect an artifact including an AI-generated video  with only using simple sanity checks and a plain 2D CNN architecture.

\section{Materials and Methods}

\subsection{Data Collection and Annotation}
To investigate artifacts in AI-generated videos, we used Sora to create 100 short video clips (each 10 seconds in length). These videos were prompted with concise phrases combining people and locations (e.g., “a high school student walking across the Shibuya crossing”). After reviewing the overall content, we selected 15 videos for more detailed analysis, extracting two frames per second from each video, resulting in 300 still images. 

We defined four artifact categories based on a preliminary survey of common AI artifacts. 

\noindent\textbf{Label 1.} Boundary / edge defects: Blurred or jagged edges separating foreground objects from the background.

\noindent\textbf{Label 2.} Texture / noise issues: Excessively uniform or noisy patches in textures (e.g., on clothing or walls).

\noindent\textbf{Label 3.} Movement/joint anomalies: Limbs refracted in opposite directions or body motility unnatural.

\noindent\textbf{Label 4.} Object mismatches / disappearances: Inconsistent appearance or disappearance of objects or body parts.

If the corresponding artifacts were present, each of the 300 images was then annotated with one or more of these four labels.

\subsection{Model Selection and Training}
We employed a 2D Convolutional Neural Network (CNN), specifically ResNet-50, EfficientNet-B3, EfficientNet-B4, and ViT-Base, which has shown effectiveness in image classification tasks. The model was adapted for multi-label classification by generating four independent outputs, one for each artifact type. The dataset was randomly divided into training (80\%) and validation (20\%) sets to assess model performance. 

\subsection{Evaluation Metrics}

To evaluate the model’s ability to detect each artifact category accurately, we calculated the following. (i) \textbf{Macro-average accuracy.} The average of the proportion of correctly predicted labels calculated separately for each class in the validation set. (ii) \textbf{Loss curves.} Tracking training and validation loss across epochs helped identify potential overfitting.  (iii) \textbf{Grad-CAM Visualization.} Grad-CAM was used to visualize which parts of the image the model focused on for each image, and the results were compared to the human-defined labels.

\section{Experiments}
\subsection{Artifact Frequency}

An initial analysis revealed the frequency of each artifact category in the 300 annotated images. As shown in Table 1, the object mismatches / disappearances category was the most prevalent, appearing in 241 images (80.3\%). Meanwhile, boundary / edge Defects were found in 134 images (44.6\%), and texture / noise issues were observed in 77 images (25.6\%). The movement / joint anomalies category had a lower incidence but was noted in 49 images (16.3\%). Since multiple artifact types can occur in the same image, the sum of percentages exceeds 100\%.

% table1 main_result
\begin{table}[t]
    \centering
    \caption{%
        Artifact Frequency in Sora-generated Video Frames. We report both the absolute frequency count and the corresponding percentage for each category. \hk{L1, 2, 3, and 4 indicates Label 1, 2, 3, and 4 in the table. Note that multiple artifact types are simultaneously occurred at the same image. This means the sum of percentages exceeds 100\%.}
    }
    \vspace{-10pt}
    \begin{tabular}{l|c|c}
    \toprule
    \textbf{Category} & \textbf{Count} & \textbf{\%} \\
    \midrule
    L1: Boundary / Edge Defects & 134 & 44.6 \\
    L2: Texture / Noise Issues & 74 & 24.6 \\
    L3: Movement / Joint Anomalies & 78 & 26.0 \\
    L4: Object Mismatches / Disappearances & 241 & 80.3 \\
    \bottomrule
    \end{tabular}
    \label{tab:main_result}
\end{table}

\subsection{Model Performance}
Our experiments revealed that ResNet-50 achieved the highest mean per label accuracy at 94.14\%, demonstrating its strong ability to capture subtle generative artifacts via residual connections and original architecture. EfficientNet-B3 followed with a mean accuracy of 93.36\%, while EfficientNet-B4 and ViT-Base recorded mean accuracies of 92.19\% and  92.97\%, respectively. EfficientNet-B4 excelled in detecting Label 2 with an accuracy of 96.88\%, but the detection accuracy for Label 3 was relatively lower at 88.75\%. The ViT-Base showed a high accuracy of 96.88\% with label 2 but with Label 3 reaching only 89.06\%. 

ResNet-50 is notable not for having outstanding values for individual labels but for consistently demonstrating high accuracy across all labels without significant drops. Specifically, although label 1 falls slightly short compared to EfficientNet-B3 or ViT-Base, it achieves a comparable 96.88 percent for label 2, the highest accuracy of 93.75 percent for label 3. It outperforms the others at 95.31 percent for label 4. In other words, it is not that one particular label shows extremely high accuracy, but overall balanced high performance across all labels that contributes to the highest average percentage of correct answers. Regarding labels 3 and 4, where ResNet-50 performed better than the other models, ResNet-50 may respond better to movement/joint anomalies and object mismatches / disappearances. However, since the study used a limited data set, the simplicity of the model structure may just have influenced the results. Conversely, there is some potential for detection even with a small number of samples, so visual artifact detection with limited data may be possible.

Figure 2 illustrates the training and validation loss curves over the epochs. While the training loss steadily declined, the validation loss exhibited minor fluctuations, indicating a degree of overfitting. Nonetheless, the model consistently recognized the most frequently occurring artifact, object mismatches / disappearances, suggesting that CNNs may better detect visually apparent discontinuities.

\begin{table*}[t]%[htbp]
    \centering
    \caption{Assessment results for each model.}
    \vspace{-10pt}
    \begin{tabular}{lccccc}
        \hline
        \textbf{Model} & \textbf{Label 1(\%)} & \textbf{Label 2(\%)} & \textbf{Label 3(\%)} & \textbf{Label 4(\%)} & \textbf{Mean(\%)} \\
        \hline
        ResNet-50         & 90.62 & \textbf{96.88} & \textbf{93.75} & \textbf{95.31} & \textbf{94.14} \\
        EfficientNet-B3 & \textbf{95.31} & 96.31 & 89.06 & 92.81 & 93.36 \\
        EfficientNet-B4 & 90.62 & \textbf{96.88} & 88.75 & 92.50 & 92.19 \\
        ViT-Base        & 93.75 & \textbf{96.88} & 89.06 & 92.19 & 92.97 \\
        \hline
    \end{tabular}
    \label{tab:mpla_results}
\end{table*}

% fig
\begin{figure}[t]
    \centering
    \includegraphics[width=1.0\linewidth]{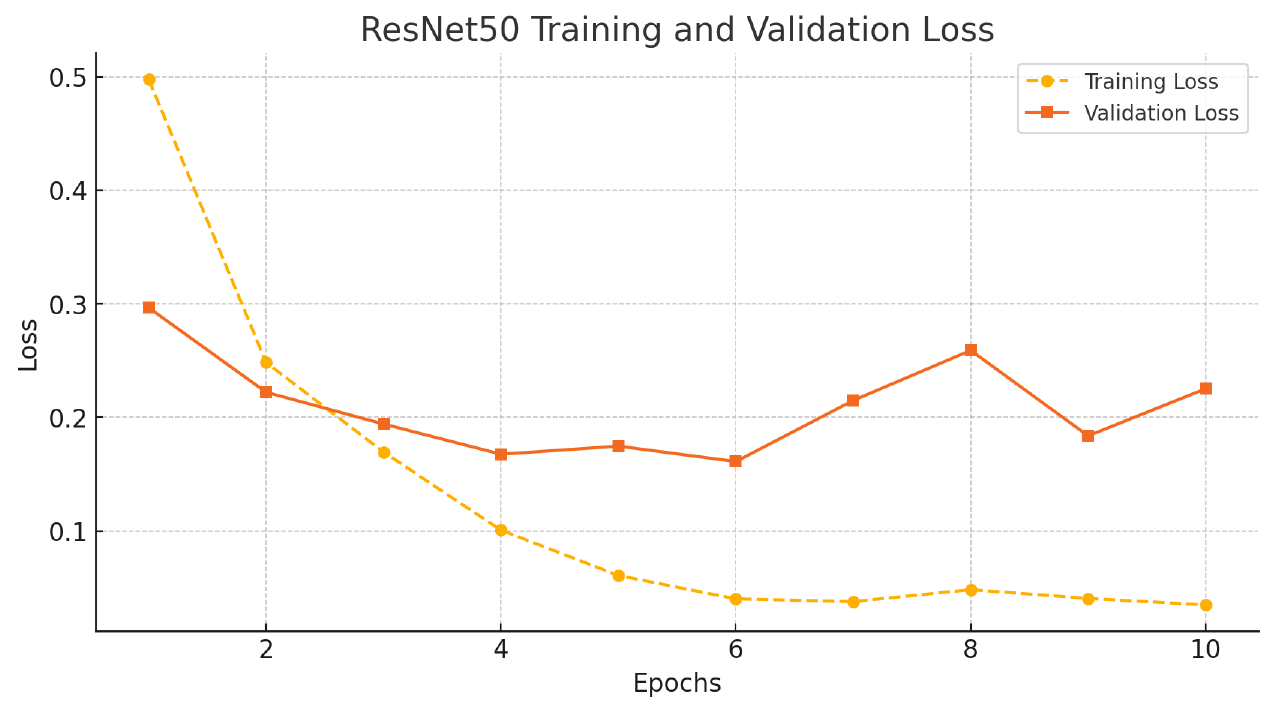}
    \vspace{-15pt}
    \caption{The training and validation loss curves.}
    \vspace{-10pt}
    \label{fig:dataset_pipline}
\end{figure}

\subsection{Grad-CAM Observations}
While recent deep learning models have shown very high performance, their decision-making process tends to be a black box. Therefore, we applied “Gradient-weighted Class Activation Mapping (Grad-CAM),” which visually explains which part of the deep neural network is focused on to make a decision~\cite{selvaraju2017grad}. Grad-CAM visualizes important regions in an image for a specific class by generating a heatmap. Grad-CAM helps us understand the model’s decision process. Warm colors (like red) indicate areas with high importance, while cool colors (like blue) show less relevant regions. Interestingly, the model sometimes focused on regions a human annotator might not consider central to the artifact. For example, humans may label frames with incorrect body position, whereas a model may concentrate on the background texture. This discrepancy highlights the possibility that AI leverages features not immediately apparent to human observers.

% fig
\begin{figure}[t]
    \centering
    \includegraphics[width=0.95\linewidth]{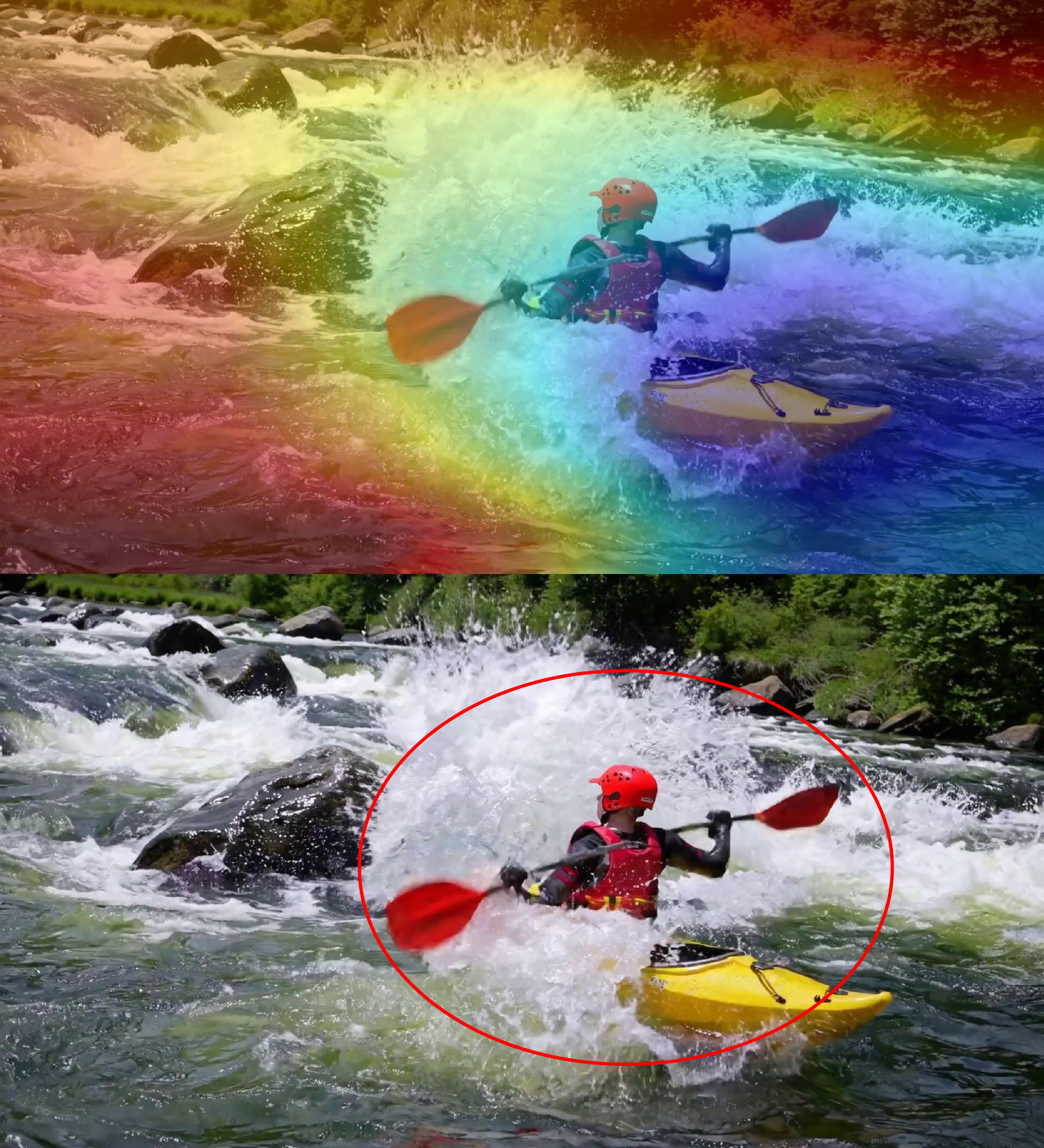}
    \vspace{-7pt}
    \caption{Difference between where the model and human focus their attention. Above: where the model makes decisions (Grad-CAM) / Below: where the human pays attention.}
    \vspace{-15pt}
    \label{fig:dataset_pipline}
\end{figure}

\subsection{Limitations}

\noindent\textbf{Limited dataset.} With only 15 thoroughly analyzed videos (300 frames), the dataset may not fully represent the breadth of artifacts possible in diverse prompts or scenes.

\noindent\textbf{Overfitting / domain shift.} Oscillating validation loss shows potential overfitting; real-world performance may differ.

\noindent\textbf{Temporal dynamics.} Analyzing frames overlooks crucial motion-based clues. Movement anomalies are inherently temporal, suggesting the potential benefit of 3D CNNs or optical flow algorithms. 

\section{Conclusion}
This study examined Sora-generated videos and identified four key artifact categories, label 1: boundary / edge defects, label 2: texture / noise issues, label 3: movement / joint anomalies, and label 4: object mismatches / disappearances. In summary, our multi-label analysis framework effectively predicts multiple defect labels on each AI-generated image, demonstrating its capability to capture a range of subtle generative artifacts. 

Among the evaluated architectures, ResNet-50 emerged as the strongest performer with a mean per-label accuracy of 94.14\%, highlighting the benefits of residual connections for extracting fine-grained features. Although variations were observed across defect types—such as EfficientNet-B4’s exceptional performance for label 3 versus its lower accuracy for label 4—these differences underscore the importance of aligning model characteristics with specific defect representations. However, the accuracy of all models exceeds 90\% so visual artifact detection with limited data may be possible.

The overall high classification scores suggest that the inherent characteristics of AI-generated imagery, including distinct artifacts produced during the generative process, play a significant role in facilitating label prediction. While the findings underscore the feasibility of automated artifact detection, several avenues for improvement remain. Increasing the dataset size, adopting temporal analysis (e.g., 3D CNNs, optical flow), and combining real and synthetic videos for training will likely enhance model robustness. 

It is noteworthy that in this work, the number of trained data was small, so the simplicity of the model structure may have only affected the high accuracy. In contrast, it can be interpreted that there is some potential for detection even with a small number of data sets, and it may be possible to say that `visual artifact detection with limited data' is feasible. Research in this direction may further enable personally customized deepfake detection with high-accuracy, a critical tool in mitigating the spread of misinformation on social media and other online platforms.

\begin{comment}
\section{References}
[1] D. M. J. Lazer, M. A. Baum, Y. Benkler, et al., “The science of fake news,” Science, vol. 359, no. 6380, pp. 1094–1096, 2018.
[2] A. Thies, M. Zollhöfer, M. Stamminger, C. Theobalt, and M. Nießner, “Face2Face: Real-Time Face Capture and Reenactment of RGB Videos,” in Proc. IEEE Conf. Computer Vision and Pattern Recognition, 2016, pp. 2387–2395.
[3] S. Suwajanakorn, S. M. Seitz, and I. Kemelmacher-Shlizerman, “Synthesizing Obama: Learning Lip Sync from Audio,” in ACM Trans. Graph., vol. 36, no. 4, Jul. 2017, pp. 1–13.
[4] I. Goodfellow, J. Pouget-Abadie, M. Mirza, B. Xu, D. Warde-Farley, S. Ozair, A. Courville, and Y. Bengio, “Generative adversarial nets,” in Advances in Neural Information Processing Systems (NIPS), 2014, pp. 2672–2680.

\end{comment}

{\small
\bibliographystyle{ieeenat_fullname}
\bibliography{main}
}

\end{document}